\documentclass[10pt,twocolumn,letterpaper]{article}

\usepackage{cvpr}              
\usepackage{algorithm}
\usepackage{algpseudocode} 
\usepackage{comment}    
\usepackage{amssymb}   
\usepackage{pifont}
\usepackage{booktabs}  
\usepackage{graphicx}  
\usepackage{multirow}
\usepackage{float}
\usepackage[accsupp]{axessibility}  

\definecolor{cvprblue}{rgb}{0.21,0.49,0.74}
\usepackage[pagebackref,breaklinks,colorlinks,allcolors=cvprblue]{hyperref}

\def\paperID{} 
\def\confName{CVPR}
\def\confYear{2026}

\title{Node-RF: Learning Generalized Continuous Space-Time Scene Dynamics\\with Neural ODE-based NeRFs}

\author{
Hiran Sarkar$^{1}$ \quad
Liming Kuang$^{1,2}$ \quad
Yordanka Velikova$^{1,2}$ \quad
Benjamin Busam$^{1,2}$ \\
$^{1}$Technical University of Munich
$^{2}$Munich Center for Machine Learning \\
{\tt\small hiransarkar2001@gmail.com, \{liming.kuang, dani.velikova, b.busam\}@tum.de}
}

\begin{document}
\makeatletter
\g@addto@macro\@maketitle{
  \vspace{-30pt}
  \begin{figure}[H]
  \setlength{\linewidth}{\textwidth}
  \setlength{\hsize}{\textwidth}
  \centering
    \includegraphics[width=\linewidth]{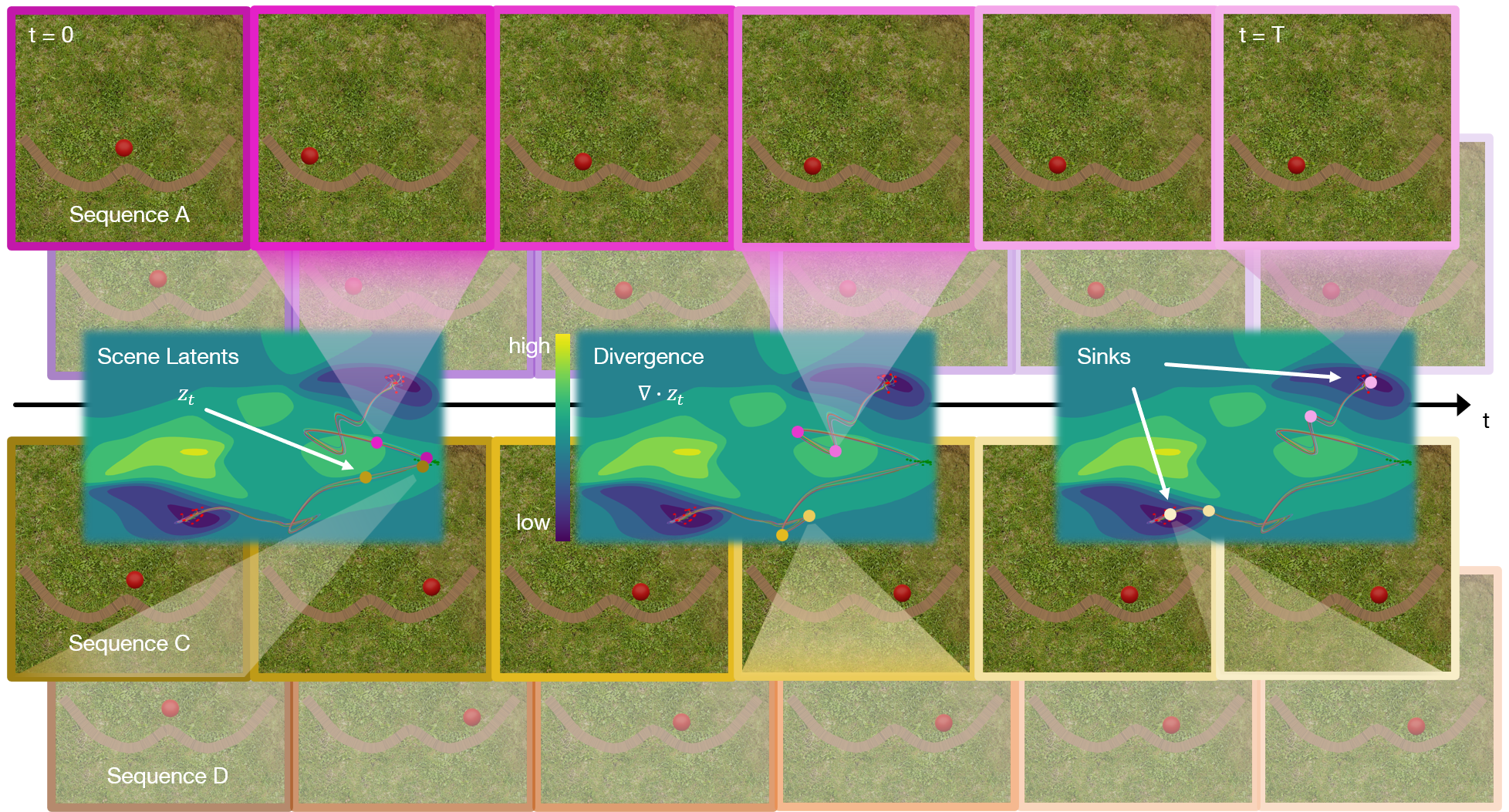}
	\caption{{\bf Node-RF Overview.} Multiple observations of the red ball being dropped onto the dual ramp end in damped oscillations towards the two valleys. Node-RF learns to encodes generalized dynamics of deterministic motions from sequences such as A (pink), C (yellow) into an implicit space-time latent representation using NeRF and nODE. Embedded sequence states depicted as scene latent points $z_t$ on the embedding space are propagated through time $t$ using an implicit neural ODE. Intermediate and future states (centre to right encodings) can be extrapolated and rendered (top, bottom) given initial frame conditions (on the left). Latent divergence $\nabla z_t$ (colour coded) characterizes behaviour of the learnt system. Various initial states of dropping the ball at locations around the centre (green) propagate through time (lines) towards the steady state sinks with $\nabla z_t < 0$ (right).
    }
    \label{fig:teaser}
	\end{figure}
}
\makeatother

\maketitle
\begin{abstract}
Predicting scene dynamics from visual observations is challenging. 
Existing methods capture dynamics only within observed boundaries failing to extrapolate far beyond the training sequence.
Node-RF (Neural ODE-based NeRF) overcomes this limitation by integrating Neural Ordinary Differential Equations (NODEs) with dynamic Neural Radiance Fields (NeRFs), enabling a continuous-time, spatiotemporal representation that generalizes beyond observed trajectories at constant memory cost.
From visual input, Node-RF learns an implicit scene state that evolves over time via an ODE solver, propagating feature embeddings via differential calculus. A NeRF-based renderer interprets calculated embeddings to synthesize arbitrary views for long-range extrapolation. 
Training on multiple motion sequences with shared dynamics allows for generalization to unseen conditions.
Our experiments demonstrate that Node-RF can characterize abstract system behavior without explicit model to identify critical points for future predictions.
Our code will be made publicly available.
\end{abstract}    
\section{Introduction}
\label{sec:intro}

Modeling the dynamics of real-world scenes from visual observations is a central problem in computer vision, with applications spanning dynamic view synthesis, robotic planning, scene forecasting, and 4D reconstruction. As cameras capture only sparse and discrete snapshots of an evolving environment, a key challenge is to infer a continuous and physically plausible representation of how the scene changes over time. Achieving this requires capturing both the underlying 3D structure and its temporal evolution in a way that generalizes across sequences even with unseen initial conditions.

Traditional approaches often rely on explicit motion models or discretized predictive architectures. Optical flow and deformation-based methods, while effective for short-term predictions, struggle with large motions, occlusions, and long-range extrapolation. Recent video prediction and generative models~\cite{dong2023video} synthesize intermediate frames at fixed time-steps.
These models struggle with irregular temporal sampling and reasoning about continuous motion over long time horizons.

\textit{Dynamic radiance fields} extend the success of Neural Radiance Fields (NeRF)~\cite{mildenhall2021nerf} to 4D spatio-temporal scenes. They learn volumetric representations that jointly encode geometry, appearance, and motion, enabling photorealistic rendering from novel viewpoints. However, most dynamic NeRF variants~\cite{pumarola2021d,park2021nerfies,lin2024dynamic} parameterize time as a discrete set of frames or rely on learned deformation fields specific to each training trajectory. This design leads to:
(1) \textbf{Limited extrapolation:} Motion is learned only at observed frames, supporting interpolation near training timestamps without principled mechanism for long-range temporal extrapolation.
(2) \textbf{Lack of generalization:} Deformation fields are sequence-specific, preventing the model to generalize to dynamic patterns that are not seen during training.

To address these challenges, we explore a complementary direction: learning continuous-time latent dynamics that explain the evolution of a 3D scene. Neural Ordinary Differential Equations (Neural ODEs)~\cite{chen2018neural} provide a framework for such a representation. They capture temporal evolution via differential operators rather than discrete transitions, enabling smooth, consistent trajectories even with irregular timestamps. While Neural ODEs have been successfully applied in dynamical systems and trajectory prediction, their integration with radiance fields for spatio-temporal reasoning remains relatively underexplored.

We introduce \textbf{Node-RF (Neural ODE meets NeRF)}, a unified framework that tightly couples Neural ODEs with a radiance field to model continuous scene dynamics directly from images. The central idea is to represent the state of a dynamic scene using a latent vector that evolves according to a learned ODE. At any queried timestamp, this latent state is decoded by a NeRF-based renderer to produce geometry and appearance that reflect the scene configuration at that moment. This design provides several advantages:

\textit{Continuous-time extrapolation:} By modeling the scene dynamics as a continuous ODE, the system can predict arbitrary timepoints while maintaining smooth, coherent evolution. This formulation reduces temporal drift and artifacts common in frame-based dynamic NeRFs and enables physically plausible \textbf{long-range extrapolation}.

\textit{Trajectory generalization:} By learning a continuous dynamic model shared across multiple trajectories governed by the same underlying physics, the system can \textbf{generalize to novel trajectories} arising from \textit{unseen initial states} while maintaining consistent continuous-time behavior.

During training, Node-RF learns from single- or multi-view image sequences using a photometric loss, without optical flow, depth, or 3D ground truth. The Neural ODE governs the temporal evolution of a canonical latent code, while the radiance field decodes the latent state for view-consistent renderings. This learning process encourages latent dynamics to reflect meaningful scene behavior, such as object motion, viewpoint variation, or non-rigid deformation.

Our main contributions are:
\begin{enumerate}
    \item Designing a novel \textbf{implicit spatio-temporal representation} by tightly integrating \textbf{Neural ODEs and NeRFs}, enabling a continuous representation for \textbf{long-term extrapolation}, and photorealistic novel view synthesis within individual sequences.
    \item Demonstrating \textbf{trajectory generalization} in dynamic scenes: by learning a continuous latent model across multiple sequences, the system can predict \textit{novel trajectories} from \textit{unseen initial conditions}.
    \item Showing how \textbf{learnt latent embeddings} can be leveraged for \textbf{generic analysis of dynamical systems} and characterization of dynamic behavior.
\end{enumerate}


\section{Related Work}
\label{sec:rw}















\textbf{Static Scene Representation}: Reconstructing static 3D scenes from 2D images is a key challenge in computer vision. Traditional methods struggle with memory efficiency and scalability. Neural implicit representations, such as NeRF ~\cite{mildenhall2021nerf}, have gained popularity by encoding scenes as continuous functions that map 3D coordinates to color and density, enabling high-quality novel view synthesis. Extensions of NeRF improve efficiency~\cite{zhao2023instant, muller2022instant, fridovich2022plenoxels} and robustness under sparse observations~\cite{jain2021putting}. Alternatively, Gaussian Splatting~\cite{kerbl20233d} uses 3D Gaussians for real-time rendering, offering better scalability and efficiency than NeRF's volumetric ray marching. However, these methods are limited in dynamic environments, prompting the need for techniques that capture both spatial and temporal changes.

\noindent\textbf{Dynamic Scene Representation}: Existing methods can be categorized into four types: warping-based, flow based, direct inference from space-time inputs, and implicit propagation methods.

\textit{Warping-based methods} model dynamic scenes by learning a time-dependent deformation field that warps a canonical space to different time instances~\cite{pumarola2021d,park2021nerfies}.
D-NeRF~\cite{pumarola2021d} and Nerfies~\cite{park2021nerfies} introduced this approach, while HyperNeRF~\cite{park2021hypernerf} extended it to higher-dimensional embeddings to handle topological scene changes.
To improve efficiency and scalability, TiNeuVox~\cite{fang2022fast} and HexPlanes~\cite{cao2023hexplane} factorize spatiotemporal features into compact voxel or plane-based structures, and K-Planes~\cite{fridovich2023k} further generalizes this via low-rank feature decompositions.
Nerfies additionally employs elastic regularization to encourage smooth, physically plausible deformations and reduce overfitting to transient motion artifacts.

\textit{Flow-based methods} explicitly estimate scene motion to guide radiance modeling.
NSFF~\cite{li2022neural} learns dense \textit{scene flow} across frames, enabling motion-aware novel view synthesis and temporally consistent reconstruction.

\textit{Direct inference methods} bypass deformation or flow estimation and directly infer scene appearance from spatiotemporal inputs.
DyNeRF~\cite{li2022neural} trains a NeRF conditioned on 3D positions and per-frame latent codes to represent time-varying radiance fields, a setup we also build upon by predicting these latent trajectories through a neural dynamical model.
Explicit representations such as MotionGS~\cite{zhu2024motiongs}, MAGS~\cite{guo2024motion}, and 4D Gaussian Splatting~\cite{wu20244d} extend 3D Gaussian Splatting~\cite{kerbl20233d} to the temporal domain, efficiently capturing space–time variations with explicit point-based primitives.

\textit{Implicit propagation methods} model temporal evolution via implicit continuous dynamics parameterization.
DONE~\cite{wang2024done} leverages neural ODE to perform dynamic reconstruction from posed multi-view images but relies on a two-stage, mesh-based pipeline that first reconstructs a static canonical mesh scaffold and then learns surface deformations. In contrast, our method directly integrates a neural ODE within a NeRF-based inverse rendering framework, learning a continuous-time implicit volumetric 3D representation directly from posed multi-view images, without mesh scaffolds or intermediate static reconstructions.
MonoNeRF~\cite{tian2023mononerf} generalizes to multiple scenes of monocular videos requiring optical flows, depth maps and binary masks as additional supervisions.


Most existing approaches learn scene-specific motions and struggle to generalize beyond the training sequence. In particular, they are not equipped to generalize to multiple sequences or learn extrapolation far beyond the training domain.
To overcome this, we augment dynamic NeRFs with the ability to learn beyond the input frames and seen dynamics by leveraging the expressive power of Neural ODEs~\cite{chen2018neural}. Our framework captures continuous-time scene dynamics, allowing for accurate temporal extrapolation and improved generalization to unseen motion patterns.

\noindent\textbf{Neural ODEs:} Neural ODEs~\cite{chen2018neural} reinterpret neural network forward passes as continuous-time dynamics by modeling hidden state evolution through ordinary differential equations. Latent ODEs~\cite{rubanova2019latent} extend this framework with an ODE-RNN to handle irregularly sampled time-series, enabling flexible continuous-time representations. Vid-ODE~\cite{park2021vid} further applies Neural ODEs to video modeling, parameterizing motion trajectories as ODE solutions for temporal interpolation and extrapolation. However, these methods remain limited to 2D temporal modeling and do not generalize to 3D dynamic scene representations. Our work extends Neural ODEs to dynamic NeRFs, enabling spatial and temporal generalization across 3D scenes.

\section{Preliminaries: Neural RF and ODE}
\label{sec:preliminaries}
\begin{figure*}[t!]
    \centering
    \includegraphics[width=1.0\textwidth]{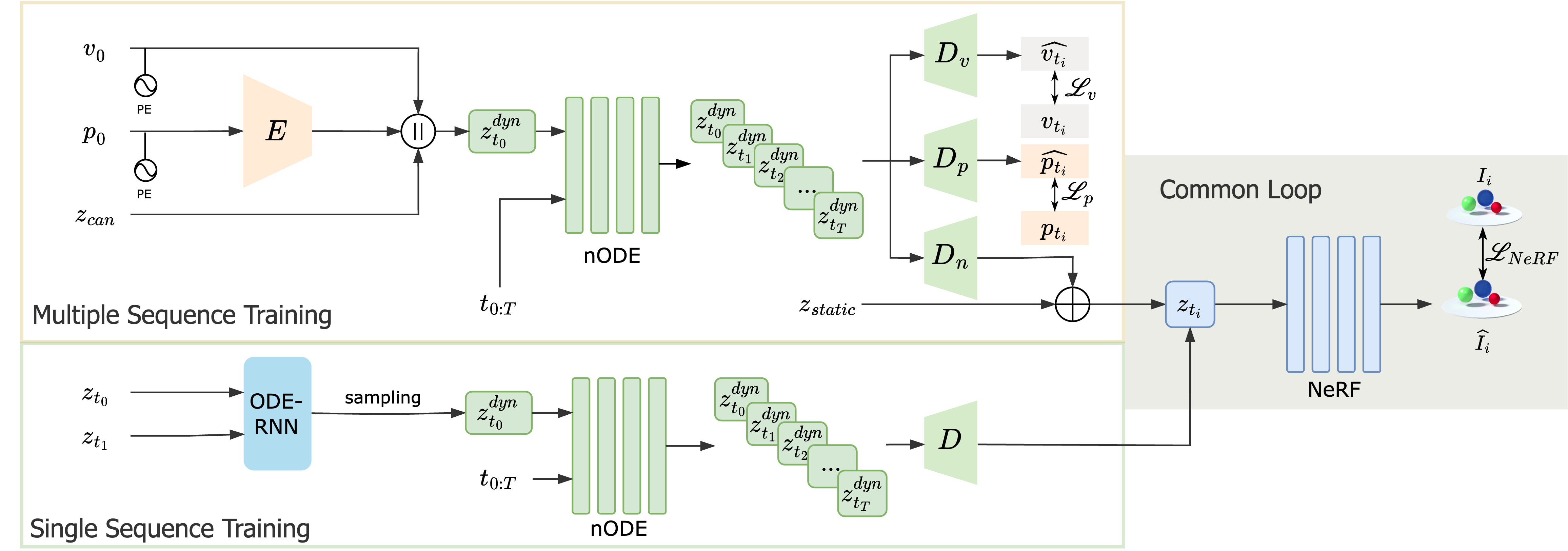}
    \caption{\textbf{Overview of Node-RF}. $z_{can}$, $z_{t_0}$, $z_{t_1}$ are learned during warmup iterations. The \textit{Multiple Sequence Training} along with \textit{NeRF} (Common Loop) illustrates the \textbf{Generalized Multi-Sequence Learning} task. The \textit{Single Sequence Training} along with \textit{NeRF} (Common Loop) illustrates the \textbf{Continuous Single-Sequence Dynamics} task.}
    \label{fig:pipeline}
\end{figure*}

\textbf{Dynamic-NeRF.}
Given a 3D position $\mathbf{x} \in \mathbb{R}^3$ and a 2D viewing direction $\mathbf{d} \in \mathbb{R}^2$, NeRF \cite{mildenhall2021nerf} aims to estimate volume density $\sigma \in \mathbb{R}$ and emitted RGB color $\mathbf{c} \in \mathbb{R}^3$ using neural networks. Following DyNeRF \cite{li2022neural}, we formulate the dynamic NeRF $F_{\Theta}$ by learning latent representation $z_t$ coresponding to each timestep such that
\begin{equation}
    F_{\Theta} \left( \mathbf{x}, \mathbf{d}, z_t \right) = (\mathbf{c}, \sigma).
\end{equation}

$F_{\Theta}$ is implemented as fully connected neural networks. From the camera with origin $\mathbf{o}$ and ray direction $\mathbf{d}$, the color of camera ray $\mathbf{r}(s) = \mathbf{o} + s\mathbf{d}$ at time frame $t$ is  
\begin{equation}
    C_t(\mathbf{r}) = \int_{s_n}^{s_f} T(s) \sigma(\mathbf{r}(s), z_t) \mathbf{c}(\mathbf{r}(s), \mathbf{d}, z_t) \, ds.
\end{equation}
where $s_n$ and $s_f$ denote the bounds of the volume depth range, and the accumulated opacity is given by  
\begin{equation}
    T(s) = \exp\left(-\int_{s_n}^{s} \sigma(\mathbf{r}(p), z_t) \, dp\right).
\end{equation}
We apply a hierarchical sampling strategy as in \cite{mildenhall2021nerf}, with stratified sampling at the coarse level followed by importance sampling at the fine level.

\noindent\textbf{Loss Function.} The network parameters $\Theta$ and the latent codes $z_t$ are simultaneously trained by minimizing the $\ell_2$-loss between the rendered colors $\hat{C}(\mathbf{r})$ and the ground truth colors $C(\mathbf{r})$, summed over all rays $\mathbf{r}$ that correspond to the image pixels from all training camera views $R$, coarse-to-fine levels $J$, and throughout all time frames $t \in T$ of the recording:  
\begin{equation}
    \mathcal{L}_{NeRF} = \sum_{t \in T,\ j \in J,\ \mathbf{r} \in R}
    \left\| \hat{C}_{t}^j(\mathbf{r}) - C_{t}(\mathbf{r}) \right\|_2^2.
    \label{eq:nerf}
\end{equation}




\noindent\textbf{Neural-ODE.}
Neural ODEs~\cite{chen2018neural} -- nODEs for short -- are a class of continuous-time models where the evolution of a hidden state \( h(t) \) is governed by an ordinary differential equation (ODE):

\begin{equation}
\frac{d h(t)}{dt} = f_{\theta}(h(t), t), \quad \text{with} \quad h(t_0) = h_0
\end{equation}

Here, \( f_{\theta} \) is a neural network parameterized by \( \theta \), which defines the dynamics of the hidden state over time. Unlike discrete models, nODEs allow the state \( h(t) \) to be continuously evaluated at arbitrary time steps using a numerical ODE solver:
\begin{equation}
h_0, \dots, h_N = \text{ODESolve}(f_{\theta}, h_0, (t_0, \dots, t_N)).
\end{equation}

\section{Methodology of Node-RF}
\label{sec:method}



We present \textit{Node-RF}, a novel framework that integrates NeRF with ODEs to model dynamic scenes. The ODE \( f_{\theta} \) captures temporal evolution in the latent space, generating a view-agnostic latent representation \( z_t \) for each timestep. This latent is then passed to the dynamic NeRF \( F_{\Theta} \), which learns the spatial characteristics of the scene. Our entire framework is trained end-to-end, enabling temporal and spatial learning from visual with:
\begin{align}
    z_{t_0}, \dots, z_{t_N} &= \text{ODESolve}(f_{\theta}, z_{t_0}, (t_0, \dots, t_N)),\\
    F_{\Theta} \left( \mathbf{x}, \mathbf{d}, z_{t_i} \right) &= (\mathbf{c}, \sigma).
\end{align}
Augmenting NeRF with ODE helps our implicit representation to learn the underlying process rather than memorizing specific states.
To demonstrate the effectiveness of our approach, we conduct two subtasks:
\begin{enumerate}
    \item We learn single-sequence dynamics from a video and validate our approach through smooth fine-grained interpolations and long-term extrapolations.
    \item We generalize scene dynamics by learning from multiple sequences governed by the same underlying motion.
\end{enumerate}

\subsection{Continuous Single-Sequence Dynamics}
In this task, we follow a similar data setup to D-NeRF \cite{pumarola2021d} and DyNeRF \cite{li2022neural}, where a dynamic scene is captured from multiple viewpoints. We leverage latent-ode \cite{rubanova2019latent} to model the temporal evolution of the scene, which uses an ODE-RNN variational autoencoder to learn the initial hidden state $z_{t_0}^{\text{dyn}}$.
We first learn the latent representations corresponding to the first two frames ($z_{t_0}$, $z_{t_1}$) in a warm-up phase. The latents codes are fed to the NeRF model $F_{\Theta}$ which learns from the first two corresponding ground-truth frames, while the nODE is kept frozen. After this, we jointly train the nODE with the NeRF. The two learned latents are fed into the ODE-RNN autoencoder, which learns a normal distribution over the latent space. Sampling from this distribution yields $z_{t_0}$ which is passed to the ODE solver. It generates the dynamic latent representation $z_{t_i}^{\text{dyn}}$ for the timestep $i$, which is fed to the decoder $\mathcal{D}$ to get the NeRF latent $z_{t_i}$ for rendering.
The nODE integrates a differential function over time enforcing smooth and consistent transitions between states.
This setup enables us to effectively capture the dynamics of the scene, facilitating not only fine-grained interpolation but also long-term extrapolation.
Joint training is done solely from visual observation via the reconstruction loss $\mathcal{L}_{NeRF}$.

\subsection{Generalized Multi-Sequence Learning}
To learn generalized continuous space-time dynamics from multiple scenes that follow the same motion pattern, we adopt our implicit scene representation to accommodate different initial conditions of a dynamic object.
Training one model on sequences of varying initial conditions enables the generation of unseen scene states. We exemplify initial conditions with object position and velocity. 
We initially learn a static latent $z_{static}$ to captures the static background during warmup when latent is trained with NeRF while the nODE is kept frozen. After warmup, the nODE learns dynamics by optimizing a canonical latent $z_{can}$ as scene reference. It takes positional encodings for the initial conditions of pose $p^c_{0}$, velocity $v^c_{0}$ at $t_0$.
The nODE propagates the canonical scene latent $z_{can}$ through time to predict scene representations at different times depending on the initial states. We pass $p^c_{0}$ through an MLP $\mathcal{E}$ and concatenate with initial velocity $v^c_0$ and canonical latent $z_{can}$. The result is fed to the nODE to predict corresponding dynamics and estimate latents $z_{t_i,c}^{\text{dyn}}$ at time $t_i$. This propagated latent is then passed through three distinct decoders. One to predict the scene dynamics and two for auxiliary supervision:
\begin{itemize}
    \item A \textbf{NeRF Decoder} $\mathcal{D}_n$ outputs a dynamic latent capturing motion dependant features which are added to the static latent $z_{static}$. The resulting latent $z_{t_i,c}$ is subsequently used from a NeRF model to render images.
    \item A \textbf{Pose Decoder} predicts object poses $\hat{p}^c_{t_i}$.
    \item A \textbf{Velocity Decoder} estimates the object velocity $\hat{v}^c_{t_i}$.
\end{itemize}
We use L1 losses to supervise pose and velocity with ground-truth data with
\begin{align}
\mathcal{L}_{\text{p}} = \frac{1}{T} \sum_{i=0}^{T} \left| \hat{p}^c_{t_i} - p^c_{t_i} \right|,\ 
\mathcal{L}_{\text{v}} = \frac{1}{T-1} \sum_{i=0}^{T-1} \left| \hat{v}^c_{t_i} - v^c_{t_i} \right|
\label{eq:auxiliary}
\end{align}
where $p^c_{t_i}$ and $v^c_{t_i}$ are the ground-truth positional encoded pose and velocity, of sequence $c$ at time $t_i$, and $T$ is the sequence length.
The rendered images are supervised with the NeRF reconstruction loss $\mathcal{L}_{NeRF}$.

\subsection{Training Objective}
Our overall training loss $\mathcal{L}$ is the weighted sum
\begin{equation}
    \mathcal{L} = \lambda_{\text{1}} \mathcal{L}_{\text{NeRF}} + \lambda_{\text{2}} \mathcal{L}_{\text{p}} + \lambda_{\text{3}} \mathcal{L}_{\text{v}} + \lambda_{\text{4}} \mathcal{L}_{\text{lipschitz}},
\end{equation}
where $\mathcal{L}_{\text{NeRF}}$ is the reconstruction loss from eq.~(\ref{eq:nerf}), the losses $\mathcal{L}_{\text{p}}$ and $\mathcal{L}_{\text{v}}$ are the auxiliary training signals for object pose and velocity from eq.~(\ref{eq:auxiliary}), and $\mathcal{L}_{\text{lipschitz}}$ is a regularizer described hereafter.

\noindent\textbf{Lipschitz Regularization.} We employ a regularization technique that penalizes the upper bound of the network's Lipschitz constant to promote smoother latent representations~\cite{liu2022learning}. This results in a more structured latent space, enhancing the generalization of dynamics across multiple sequences. We apply this regularization on the NeRF linear layers, where we augment each layer $y = \sigma(W_i x + b_i)$
 with a Lipschitz weight normalization layer. Utilizing a trainable Lipschitz bound $c_i$ in layer $i$, we inject
\begin{align}
    {W}_i \gets \text{normalization}(W_i, \text{softplus}(c_i)),
\end{align}
where the trainable Lipschitz bounds $c_i$ in layer $i$ bounds are optimized using
\begin{align}
    \textstyle
    \mathcal{L}_{\text{lipschitz}} = \prod_{i} \text{softplus}(c_i).
\end{align}
\section{Experimental Evaluation}
\label{sec:experiments}
\textbf{Implementation Details. }
We implement our approach in PyTorch, using 512-dimensional latent codes. Latents are initialized as \(\mathcal{N}(0, \sqrt{0.01} D)\) with \(D = 512\). Optimization employs Adam (\(\beta_1 = 0.9\), \(\beta_2 = 0.999\), learning rate \(5e-4\)). We use the \texttt{dopri5} solver for the \textit{Bouncing Balls} dataset and the \texttt{Euler} solver for the others. For \textit{Bouncing Balls}, \texttt{dopri5} provides greater stability in long-term extrapolation compared to \texttt{Euler}, whereas for the remaining datasets, \texttt{Euler} suffices, with no significant improvements observed from \texttt{dopri5}. We use the following tolerance values for nODE, atol=1e-3 and rtol=1e-4 for all our experiments, with step-size=0.05 for \texttt{Euler} solver. 
The pose encoder has 8 linear layers with 256 hidden units, while the NeRF, pose, and velocity decoders have 3 layers with 256 units. We set \(\lambda_1 = 1\), \(\lambda_2 = \lambda_3 = 10^{-2}\), and \(\lambda_4 = 10^{-22}\). Training runs for 300k iterations (for \textit{Bouncing Balls} and \textit{Pendulum} datasets) and 500k iterations (for \textit{Sear Steak}, \textit{Oscillating Ball} and \textit{Bifurcating Hill} datasets) with warm-up phases at 5k and every 4k iterations for 200 steps. 

\begin{table}[t]
\caption{
Long-term (4$\times$) Extrapolation comparison on \textit{Bouncing Balls}~\cite{pumarola2021d}.
* indicates retraining. SIM1: \textit{X-CLIP Similarity}, SIM2: \textit{LLaVA-Video Similarity}. MS: \textit{Motion Smoothness}, SC: \textit{Subject Consistency}.
}
\vspace{-10pt}
\centering
\small
\resizebox{\columnwidth}{!}{
\begin{tabular}{l|cc|cc}
    \toprule
    \multirow{2}{*}{Method} & \multicolumn{2}{c|}{Prompt-based Evaluation} & \multicolumn{2}{c}{VBench Metrics} \\
    \cmidrule(lr){2-3} \cmidrule(lr){4-5}
     & Sim1$\uparrow$ & Sim2$\uparrow$ & MS$\uparrow$ & SC $\uparrow$ \\
    \midrule
    D-NeRF* & 0.1691 & 0.7807 & 0.99473 & 0.97352 \\
    4D-GS* & 0.1484 & 0.7230 & 0.99538 & 0.92589 \\
    HexPlane* & 0.1732 & 0.6673 & \underline{0.99617} & 0.77407 \\
    TiNeuVox* & \underline{0.1773} & \underline{0.7883} & 0.99468 & 0.96428 \\
    Motion-GS* & 0.1760 & 0.7693 & 0.99465 & \underline{0.97562} \\
    Ours* & \textbf{0.1775} & \textbf{0.7937} & \textbf{0.99648} & \textbf{0.97775} \\
    \bottomrule
\end{tabular}
}
\label{tab:bb_4x_extrap}
\end{table}

\begin{table}[t]
\caption{Comparison on \textit{Pendulum}~\cite{hofherr2023neural}. Evaluated only on the foreground dynamic part. Interpolation and extrapolation are reported. A \ding{51} under TC i.e. \textit{Temporal Continuity} indicates that the method models continuous temporal dynamics.}
\vspace{-10pt}
\small
\centering
\resizebox{\columnwidth}{!}{
\begin{tabular}{l|c|ccc|ccc}
    \toprule
    Pendulum & TC & \multicolumn{3}{c|}{Interpolation} & \multicolumn{3}{c}{Extrapolation} \\
    \midrule
    Method &  & SSIM$\uparrow$ & LPIPS$\downarrow$ & PSNR$\uparrow$ & SSIM$\uparrow$ & LPIPS$\downarrow$ & PSNR$\uparrow$ \\
    \midrule
    SimVP & \ding{55} & - & - & - & \textbf{0.617} & \textbf{0.0194} & \underline{15.804} \\
    D-NeRF & \ding{51} & 0.437 & 0.0333 & \underline{13.906} & 0.426 & 0.0374 & 13.295 \\
    4D-GS & \ding{51} & \underline{0.455} & \underline{0.0300} & 13.391 & 0.463 & 0.0310 & 12.940  \\
    Ours & \ding{51} & \textbf{0.531} & \textbf{0.0234} & \textbf{17.057} & \underline{0.469} & \underline{0.0257} & \textbf{15.920} \\
    \bottomrule
\end{tabular}
}
\label{tab:pendulum_comparison}
\end{table}

\begin{table}[t]
\caption{Comparison on \textit{Oscillating Ball} (3D) and \textit{Bifurcating Hill} (2D) on IoU between GT and prediction flows. 
A \ding{51} under \textit{3D} indicates that the method can handle 3D scenes.}
\vspace{-10pt}
\small
\centering
\resizebox{\linewidth}{!}{
\begin{tabular}{l|c|cc}
    \toprule
    Method & 3D & Oscillating Ball & Bifurcating Hill \\
    \midrule
    Vid-ODE & \ding{55} & - & 0.000 \\
    SimVP & \ding{55} & - & \underline{0.295} \\
    D-NeRF(c) & \ding{51} & 0.0008 & 0.003 \\
    Ours & \ding{51} & \textbf{0.3327} & \textbf{0.485} \\
    \bottomrule
\end{tabular}
}
\label{tab:comparison_multiseq}
\end{table}

\noindent\textbf{Datasets.}
We use different sets of datasets for \textit{single sequence} and \textit{multi-sequence generalization} experiments.
\noindent\textbf{Single Sequence.}
We evaluate on the following datasets:
(1)~\textbf{Bouncing Balls}~\cite{pumarola2021d}, a deterministic scene with three bouncing balls (150 frames). 
(2)~\textbf{Pendulum}~\cite{hofherr2023neural}, a deterministic 100-frame damped pendulum sequence from a fixed viewpoint. Training is performed on alternate frames, with the last five frames reserved for extrapolation. A static background latent \( z_{static} \) is learned separately.
(3)~\textbf{DyNeRF}~\cite{li2022neural}, a real-world non-deterministic multi-view dataset captured with 21 synchronized GoPro cameras. We train on 18 cams and provide the qualitative results on the 19th cam.
\noindent\textbf{Multiple sequences.}
We develop two deterministic synthetic datasets with static background:
(1)~\textbf{Oscillating Ball}: A multi-view dataset where a ball oscillates in a bowl, captured by 9 static cameras capturing 17 sequence of varied initial ball position and zero velocity. We train on 16 sequences, reserving an unseen one for evaluation.
(2) \textbf{Bifurcating Hill}: A single-view dataset where a ball rolls down a hill and settles in one of two troughs. We train on 8 sequences and evaluate on the remaining one. Ground-truth 2D poses are obtained via CNOS~\cite{nguyen2023cnos}.

Further details on training/testing splits and dataset configurations are provided in the supplementary material.

\noindent\textbf{Baselines and Evaluation Metrics.}
For the single-sequence reconstruction and extrapolation tasks, we compare our approach against: D-NeRF~\cite{pumarola2021d} (NeRF-based scene reconstruction), 4D-GS~\cite{wu20244d} (Gaussian Splatting-based reconstruction), MotionGS~\cite{zhu2024motiongs} (flow-based reconstruction), HexPlane~\cite{cao2023hexplane} (feature plane-based reconstruction), TiNeuVox~\cite{fang2022fast} (voxel-based reconstruction), and SimVP~\cite{gao2022simvp} (future frame prediction).

In the generalization setup, we introduce three adaptations:  
(1) \textbf{D-NeRF(c)}: A modified D-NeRF conditioned on initial states. We evaluate this model on \textit{Oscillating Ball} (multi-view) and \textit{Bifurcating Hill} (single-view).  
(2) \textbf{SimVP}~\cite{gao2022simvp}: A CNN-based video prediction model limited to single-view setups, evaluated on \textit{Bifurcating Hill}.  
(3) \textbf{Vid-ODE}~\cite{park2021vid}: A neural ODE-based video generation model limited to single-view setups, also evaluated on \textit{Bifurcating Hill}.

We provide further details about the baselines in the supplementary material.

\noindent\textbf{Evaluation Metrics.}
Following novel view synthesis literature~\cite{mildenhall2021nerf,kerbl20233d}, we evaluate PSNR, SSIM, LPIPS scores. We also compute similarity scores between textual prompts and visual outputs using the X-CLIP~\cite{ma2022x} and LLaVA-Video~\cite{zhang2024video} models, along with \textit{motion smoothness} and \textit{subject consistency} metrics from the VBench benchmark~\cite{huang2024vbench}.

Additionally, we evaluate dynamic prediction quality using the \textit{IoU} score from flow masks in the multi-sequence generalization task. Following~\cite{teed2020raft}, we obtain flow masks on predicted videos by binarizing the flow magnitude with a threshold $\alpha$, resulting in $\text{Mask}_{\text{GT}}(\alpha)$ and $\text{Mask}_{\text{Pred}}(\alpha)$, and report the resulting $\text{IoU}(\alpha)$ between these masks. We set $\alpha = 0.3$ for \textit{Oscillating Ball} and $\alpha = 0.5$ for \textit{Bifurcating Hill}, adapting to dataset dynamics.

\subsection{Analysis of Results}
For all tables, the best performing results are noted in \textbf{bold}, and the second best are \underline{underlined}. 

\begin{figure}
    \centering
    \includegraphics[width=1.0\linewidth]{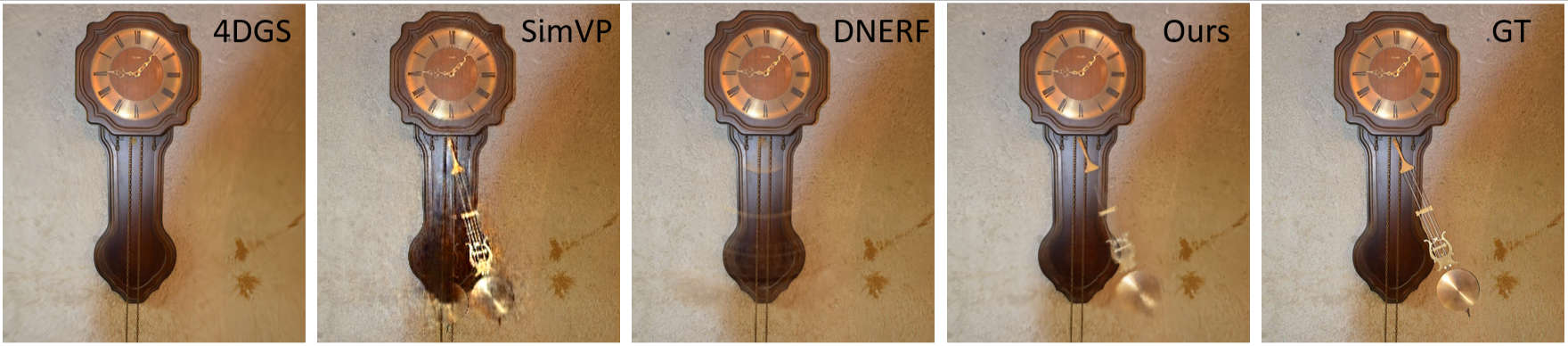}
    \vspace{-20pt}
    \caption{Extrapolation results on the \textit{Pendulum} dataset \cite{hofherr2023neural}.} 
    \label{fig:pendulum}
\end{figure}

\noindent\textbf{Long-term Extrapolation:}
The major advantage of nODEs is their ability to learn stable dynamics for long-term extrapolation. We analyze this property through both quantitative and qualitative experiments on the \textit{Bouncing Balls} deterministic scene. Table~\ref{tab:bb_4x_extrap} compares long-term extrapolation on two types of benchmarks: (1) prompt based evaluation (2) VBench metrics. 
For the prompt-based evaluation, we compute the similarity between the extrapolated videos and the following prompt:
\textit{“Does this video show physically accurate motion where the bounce amplitude gradually decreases with no sudden jumps, and the white ground plane remains completely rigid and motionless without any shaking, wobbling, or deformation?”}
Our method achieves the highest similarity across both X-CLIP and LLaVA-Video models, indicating that the generated motion adheres more closely to physically plausible dynamics.
On VBench, our approach again attains the best scores in both \textit{motion smoothness} and \textit{subject consistency}. High \textit{motion smoothness} indicates that the predicted trajectories evolve gradually over time, without abrupt jumps or jitter, reflecting physically plausible motion. Similarly, high \textit{subject consistency} ensures that individual objects retain their identity, appearance, and spatial coherence throughout the video, demonstrating that the model preserves scene integrity and perceptual coherence. Together, these results confirm that our method produces temporally stable and realistic dynamics, even over extended extrapolation horizons.

\begin{figure}
    \centering
    \includegraphics[width=1.0\linewidth]{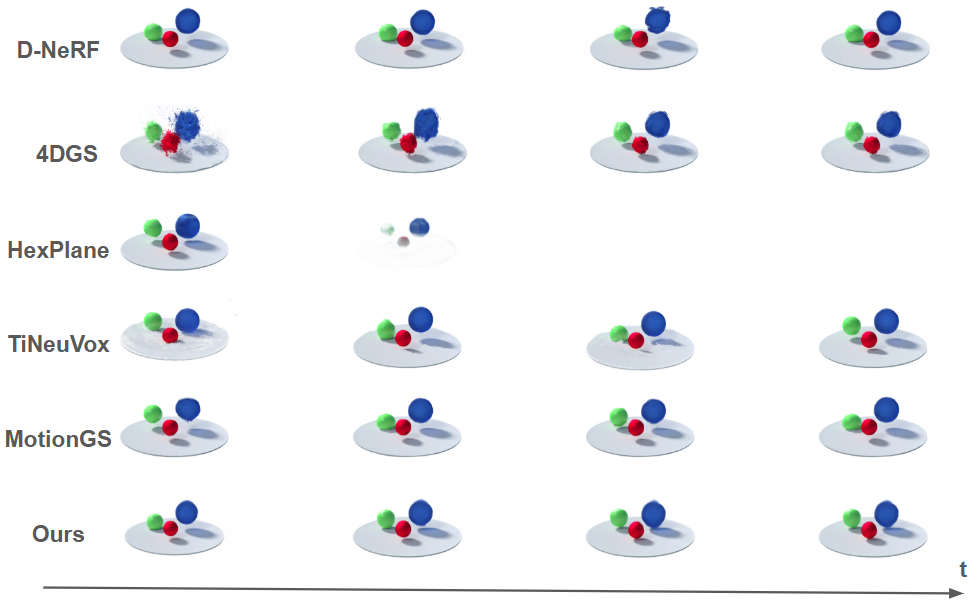}
    \vspace{-20pt}
    \caption{Long-term extrapolations, \textit{Bouncing Balls}\cite{pumarola2021d} scene.} 
    \label{fig:bb_extrapolation}
\end{figure}

Qualitative results in Fig.~\ref{fig:bb_extrapolation} further support these findings. While HexPlane causes objects to disappear, TiNeuVox, MotionGS, and D-NeRF exhibit jitter and unstable dynamics, and 4DGS results in complete disintegration of the balls, our model produces stable, physically plausible extrapolations with the balls naturally coming to rest.

\noindent\textbf{Pendulum:}
We compare our results on the \textit{Pendulum} dataset against D-NeRF, 4DGS, and SimVP. We report quantitative results in Table~\ref{tab:pendulum_comparison} on the relevant masked dynamic foreground. Both D-NeRF and 4DGS fail to capture the moving pendulum, learning only the static background. SimVP, being a frame-based future prediction model, learns a discrete representation of motion and can only extrapolate at fixed time intervals $\Delta t$, without supporting interpolation. Our method outperforms D-NeRF and 4DGS in both interpolation and short-term extrapolation, and achieves higher PSNR than SimVP during extrapolation. Qualitative results in Fig.~\ref{fig:pendulum} further show that SimVP fails to accurately capture the pendulum’s dynamics.
\noindent\textbf{Generalization:}
Our framework enables NeRF to generalize across multiple sequences governed by shared dynamic principles. Table~\ref{tab:comparison_multiseq} reports the quantitative results using the IoU metric, which measures the overlap between the predicted and ground-truth flow fields, reflecting the accuracy of dynamic modeling.  

\begin{figure}
    \centering
    \includegraphics[width=1.0\linewidth]{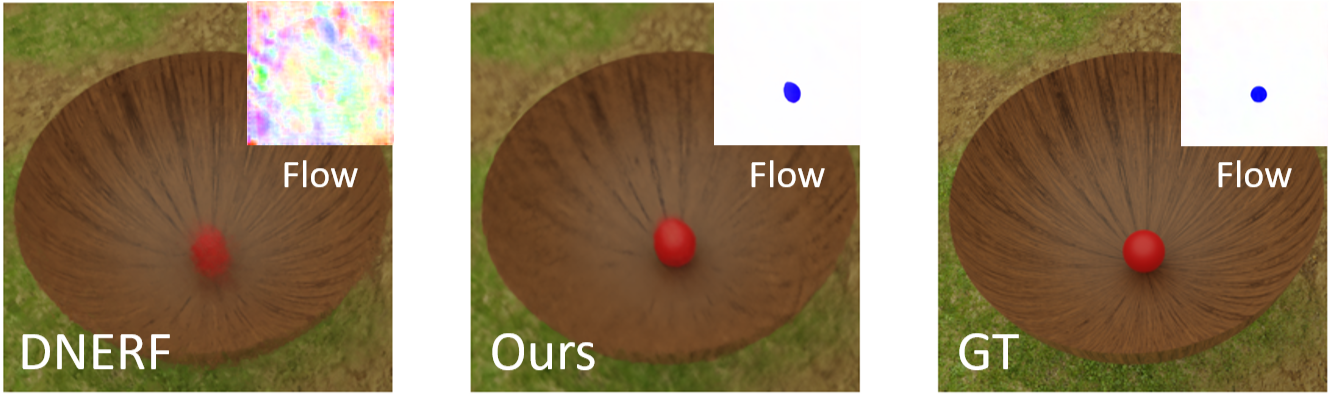}
    \vspace{-20pt}
    \caption{Dynamic Flow Masks for the predictions of a novel sequence in the \textit{Oscillating Ball} scene}  
    \label{fig:raft_bowl}
\end{figure}

\begin{figure}
    \centering
    \includegraphics[width=1.0\linewidth]{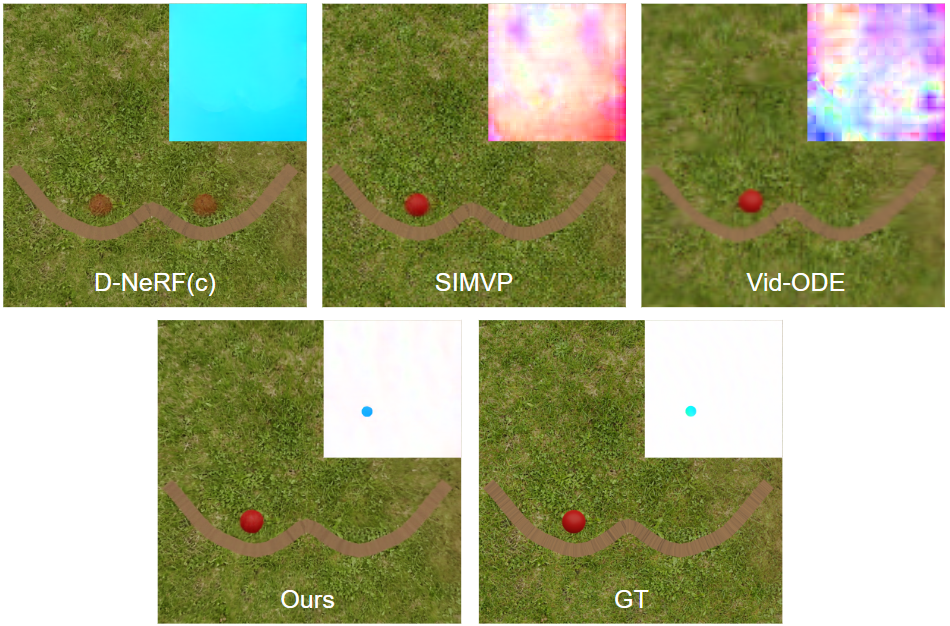}
    \vspace{-20pt}
    \caption{Dynamic Flow Masks for the predictions of a novel sequence in the \textit{Bifurcating Hill} scene} 
    \label{fig:raft_hill}
\end{figure}

For the \textit{Oscillating Ball} scene, D-NeRF(c) fails to generalize across sequences and primarily reconstructs the static background rather than the motion, resulting in significantly lower IoU scores. As shown in Fig.~\ref{fig:raft_bowl}, our method produces flow predictions closely aligned with the ground truth, whereas D-NeRF(c) yields noisy and inconsistent flow estimates. Similarly, for the \textit{Bifurcating Hill} scene, both D-NeRF(c) and Vid-ODE fail to capture the underlying dynamics, focusing instead on static regions. SimVP can predict future frames given the first two, but since it operates on fixed intervals $\Delta t$, it cannot model continuous-time dynamics. As illustrated in Fig.~\ref{fig:raft_hill}, our method achieves the highest IoU across all settings, demonstrating its superior ability to model and generalize dynamic behavior from unseen initial conditions.

\begin{figure}
    \centering
    \includegraphics[width=1.0\linewidth]{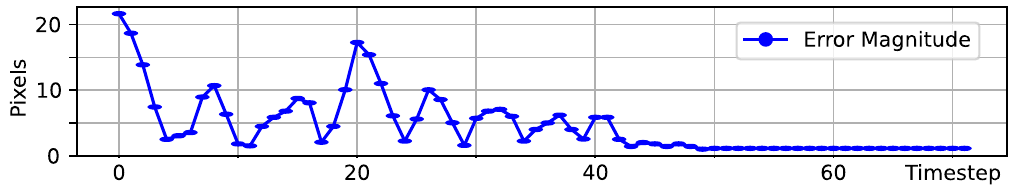}
    \vspace{-20pt}
    \caption{Pose error for a novel trajectory in \textit{Bifurcating Hill}.} 
    \label{fig:pose_error}
\end{figure}

We also provide the error plot of our predicted pose of the dynamic object (ball) of the bifurcating hill scene, in Fig.~\ref{fig:pose_error}. We utilize YOLO-v5~\cite{Jocher_YOLOv5_by_Ultralytics_2020} to localize the ball both from the predicted and the ground-truth frames, and calculate the pixel error between corresponding locations. The plot shows that, for the novel trajectory, our predictions closely follow the ground-truth motion, highlighting strong generalization and temporal consistency.

\subsection{Ablation}

\noindent\textbf{Effects of the Loss terms.}
We present the ablation results on the effect of all the loss terms in \( \mathcal{L} \) in Table~\ref{tab:ablation_losses}. When using only the NeRF reconstruction loss \( \mathcal{L}_{\text{NeRF}} \), our framework can generalize to novel initial conditions and learn meaningful dynamics. This relieves the need for the availability of ground-truth poses of the entire sequences, except for the initial conditions.
However, incorporating \( \mathcal{L}_{\text{p}} \) and \( \mathcal{L}_{\text{v}} \) further enhances the learning of dynamics. On the other hand, \( \mathcal{L}_{\text{lipschitz}} \) has a negligible effect on the quantitative results but contributes to learning a more structured latent space, as illustrated in Fig.~\ref{fig:Latent Space}. Without Lipschitz regularization, the latent codes fail to converge to a latent structure representative of the system dynamics. With the regularizer, a divergence analysis identifies steady states as low-divergence sinks as discussed in Sec.~\ref{subsec:dynamic_nerf}.

\noindent\textbf{Effects of the Latent Size}
Table~\ref{tab:ablation_latent_size} presents an ablation study on the effect of latent dimensionality, evaluated on the \textit{reconstruction} split of the \textit{Bouncing Balls} dataset. We observe that a latent size of 512 yields the best reconstruction performance. For a fair comparison, all configurations are trained for 300k iterations under identical settings.

\begin{table}[ht]
\caption{Ablation study on loss term in \textit{Oscillating Ball}.}
    \vspace{-10pt}
    \small
    \centering
    \resizebox{\linewidth}{!}{ 
    \begin{tabular}{p{3.5cm}|c c c c}
        \toprule
        {Losses Added} & {SSIM} & {LPIPS} & {PSNR} & {IoU} \\
        \midrule
        $\mathcal{L}_{\text{NeRF}}$ &  0.630 & 0.4920  &  28.661 & 0.2730 \\
        $\mathcal{L}_{\text{NeRF}} + \mathcal{L}_{\text{p}} + \mathcal{L}_{\text{v}}$ & 0.661  & 0.4396 & 29.080  & 0.3253  \\
        $\mathcal{L}_{\text{NeRF}} + \mathcal{L}_{\text{p}} + \mathcal{L}_{\text{v}} + \mathcal{L}_{\text{lipschitz}}$ & 0.662 & 0.4364 & 29.091 & 0.3327 \\
        \bottomrule
    \end{tabular}
    }
    \label{tab:ablation_losses}
\end{table}

\begin{table}[ht]
\caption{Ablation studies of the latent size on \textit{Reconstruction} set of \textit{Bouncing Balls}}
    \vspace{-10pt}
    \small
    \centering
    \begin{tabular}{c|c c c c}
        \toprule
        {\# Latent Size} & {SSIM} & {LPIPS} & {PSNR}  \\
        \midrule
        256 & 0.976 & 0.0318 & 32.29   \\
        512 & 0.978 & 0.0310 & 33.70  \\
        1024 & 0.975 & 0.0397 & 32.74  \\
        \bottomrule
    \end{tabular}
    \label{tab:ablation_latent_size}
\end{table}

\subsection{Analysis of the latent space}
\label{subsec:dynamic_nerf}

\begin{figure}
    \centering
    \includegraphics[width=1.0\linewidth]{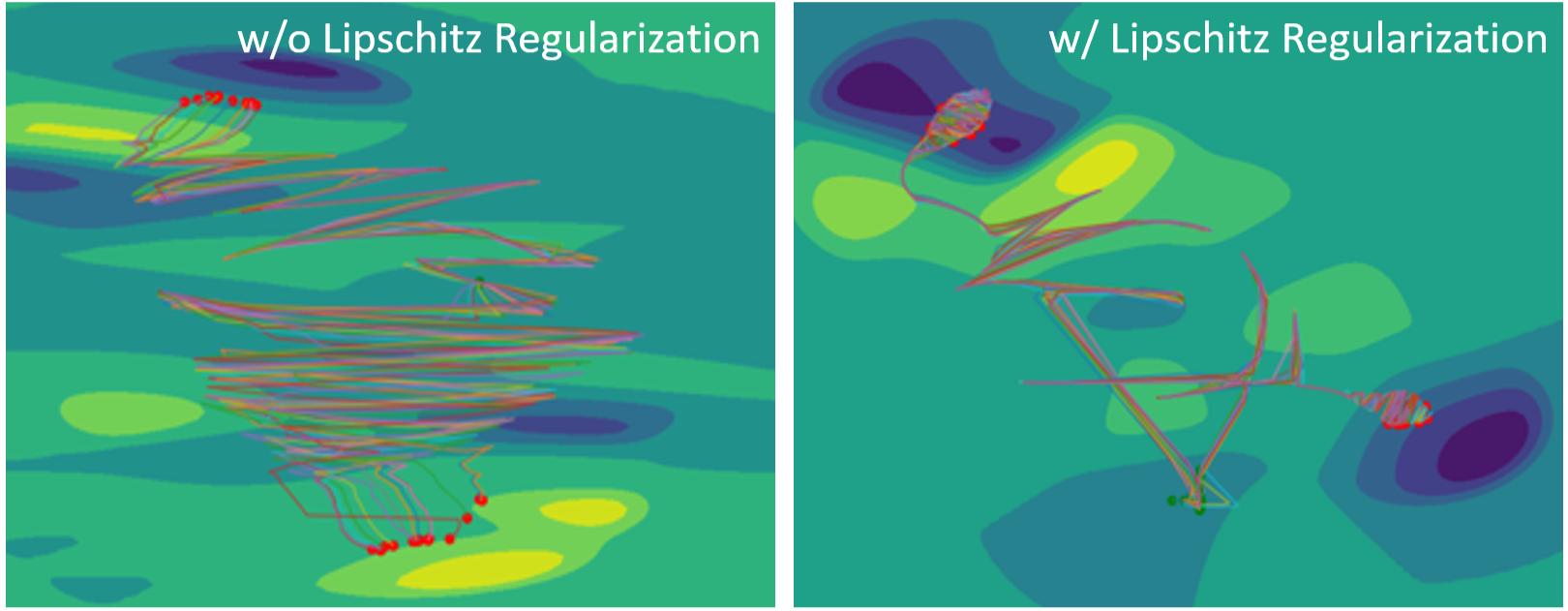}
    \vspace{-20pt}
    \caption{{Latent Space Comparison of models w/o and w/ Lipschitz regularization}. The \textcolor{green!60!black}{green} and \textcolor{red!70!black}{red} dots represent the starting and ending points of the trajectories respectively} 
    \label{fig:Latent Space}
\end{figure}

In Fig.~\ref{fig:Latent Space}, we visualize the latent space of the \textit{Bifurcating Hill} dataset. The latent representation clearly captures the dynamics of the scene, with a distinct bifurcation point (green dots) corresponding to the top of the hill. From this point, the trajectories diverge into two sinks (red dots) with low divergence, representing the ball rolling down and coming to rest in the sinks. This demonstrates that our framework effectively learns the underlying structure of the dynamical system. It allows making structural predictions on system behaviour by latent space analysis. On top of that, representative elements and trajectories can be further visualized. For instance, a latent sampling around states close to the center hill top identify sensitivity to this initial condition by following divergent extrapolation paths of the ODE, which ultimately land in one of the sinks. This allows us to make predictions about future states and can pave the way to uncertainty estimates for ambiguous visual extrapolations.

\begin{figure}
    \centering
    \includegraphics[width=1.0\linewidth]{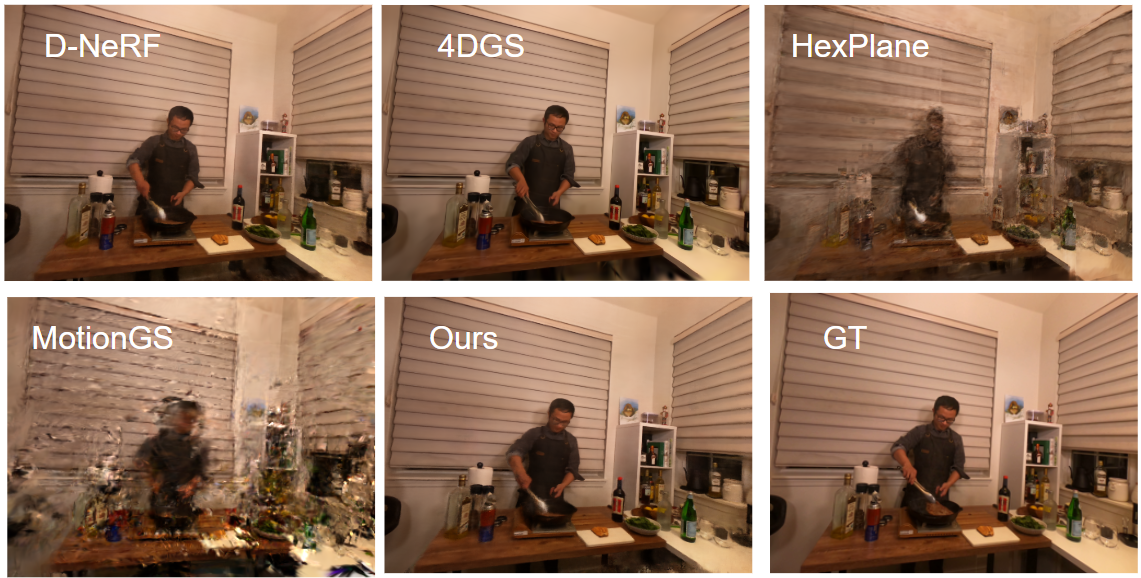}
    \vspace{-20pt}
    \caption{Extrapolation results on \textit{Sear Steak} \cite{chen2018neural}.}
    \label{fig:sear_steak}
\end{figure}

\subsection{Additional results}
\noindent\textbf{Non-deterministic dynamics:} Although our method is fundamentally designed for deterministic scene dynamics and performs best in this regime, we additionally report qualitative results on a non-deterministic scene. In this setting, HexPlane and MotionGS break down during extrapolation, while D-NeRF, 4DGS, and our approach still extrapolate reasonably, albeit with slight artifacts (Fig.~\ref{fig:sear_steak}). Even though the stochastic nature of the scene violates our modeling assumptions, our model maintains a stable temporal evolution and achieves performance on par with D-NeRF and 4DGS, indicating that it degrades gracefully under non-deterministic dynamics.
\section{Conclusion \& Limitations}
\label{sec:conclusion}
We introduced Node-RF, a novel framework that integrates Neural ODEs with NeRFs to learn continuous space-time representations of dynamic scenes. By modeling scene dynamics as an implicit function that evolves over time, our approach enables a continuous representation targeted for long-range extrapolation, and novel view synthesis. 
Unlike existing methods that rely on discretized time steps, Node-RF introduces a continuous-time framework that learns scene dynamics directly from visual observations, enabling long-horizon extrapolation without explicit physics priors.

Our experiments demonstrate Node-RF’s ability to capture complex motion patterns and generalize across scenarios, showing its potential for scientific reasoning and dynamic system analysis. However, our current results are still conceptual, serving as an early validation of the idea. We believe that our design can inspire future work, focusing on scaling to real-world datasets, improving efficiency, and extending the framework to more challenging environments. The conceptual idea of coupling an implicit dynamic reasoning model with a data-driven 3D scene representation to generate grounded 4D dynamics for behavioral system analysis goes far beyond the use of NeRFs and nODEs and may spark orthogonal ideas in the community. We believe Node-RF offers a promising step towards learning generalizable space-time representations where generation is dynamics grounded.
{
    \small
    \bibliographystyle{ieeenat_fullname}
    \bibliography{main}
}

\clearpage
\setcounter{page}{1}
\maketitlesupplementary

\section{Datasets}

\subsection{Single Sequence}

\begin{enumerate}


    \item \textbf{Pendulum:} This synthetic dataset from Hofherr et al.~\cite{hofherr2023neural} consists of 100 frames depicting a damped pendulum motion from a fixed viewpoint. Following the previous setup, we use alternate frames (0, 2, 4, ... 94) for training and the remaining frames (1, 3, 5, ... 93) for interpolation evaluation, while the last five frames (95–99) are used to assess extrapolation performance. For scene reconstruction-based methods, we select a fixed camera pose from those provided in the D-NeRF dataset. Since SimVP~\cite{gao2022simvp} assumes a fixed $\Delta t$, it cannot perform interpolation; thus, we train it on all frames up to 94 (0–94) and evaluate on the remaining five frames (95–99) for extrapolation. Our framework leverages the static background provided by the dataset by learning a separate latent code \( z_{static} \) during the warmup stage, which is added to the decoder output to obtain the NeRF latents \( z_{t_i} \)


\end{enumerate}

\subsection{Multi Sequence Generalization}
\begin{enumerate}

    \item \textbf{Oscillating Ball:} This is a multi-view synthetic dataset featuring a ball dropped into a bowl, oscillating back and forth twice. The scene is captured using nine static cameras arranged in a $3 \times 3$ grid. Each camera records two sequences with different initial positions of the ball, resulting in distinct trajectories. In total, the dataset contains 17 sequences, each corresponding to a unique initial position. All sequences start with zero initial velocity and consist of 90 frames. In addition to the multi-view images and camera poses, the dataset provides the 3D position of the ball at every timestep and an image of the static background.

    For our experiments, we subsample 25 sub-sequences from each original sequence, using the first 25 frames as the starting state. This enables the generation of sequences with varying initial velocities, computed as the difference between consecutive positions ($v_t = p_{t+1} - p_t$). We train on 16 of the original sequences, resulting in $25 \times 16$ sub-sequences with diverse initial conditions of pose and velocity. The 17$^{\text{th}}$ sequence is reserved for evaluation.

    \item \textbf{Bifurcating Hill:} This is a single-view synthetic dataset depicting a ball placed at the crest of a hill that eventually settles into a trough on either side. The dataset contains nine sequences, each consisting of 90 states. Similar to the previous setup, we subsample 25 subsequences from each original sequence, resulting in trajectories with varying initial poses and velocities. We use eight of the original sequences for training and reserve the remaining one for evaluation. For this dataset, we obtain the ground-truth 2D poses using CNOS~\cite{nguyen2023cnos}. The dataset also includes an image of the static background.

\end{enumerate}

\section{Baselines}
\subsection{Multi-Sequence Generalization}
We provide more information on the choice and implementation of the baselines used for the multi-sequence generalization experiments.

\begin{enumerate}

    \item \textbf{D-NeRF(c):} This is a modified version of D-NeRF that we adapt to incorporate conditioning on the initial conditions. Specifically, we condition the time-deformation network on the initial state, enabling it to deform rays from the frame space to a canonical space that corresponds not only to the timestep but also to the initial conditions. This allows the model to represent multiple sequences within a single implicit representation. Formally,  
    \begin{equation}
    M : (\Psi_{t,c} : (x, t, c), d) \rightarrow (c, \sigma),
    \end{equation}
    where $\Psi_{t,c}$ denotes the deformation network that warps the 3D point $x$ from the frame space to the canonical space given time $t$ and initial condition $c$. This baseline is evaluated on both the multi-view \textit{Oscillating Ball} dataset and the single-view \textit{Bifurcating Hill} dataset.


    \item \textbf{SimVP}~\cite{gao2022simvp}: This is a video prediction model built entirely on a simplified CNN architecture. We include this baseline to compare our approach against a future frame prediction model in the generalization task, as it aligns with our objective. However, SimVP is restricted to videos captured from a fixed camera viewpoint and is therefore not applicable to 3D or multi-view datasets. Consequently, we evaluate this model only on the \textit{Bifurcating Hill} dataset. For a fair comparison, we provide the first two frames as input and allow the model to predict the subsequent frames. These initial frames supply information about the starting position and velocity, ensuring consistency with our 
    framework.


    \item \textbf{Vid-ODE}~\cite{park2021vid}: This is a video generation model based on neural ODEs, designed to learn continuous-time dynamics. Since it is not equipped to handle camera poses, we evaluate it only on the 2D \textit{Bifurcating Hill} dataset. The model takes two input frames and predicts the subsequent two frames, ensuring a fair comparison with our method. During inference, it autoregressively generates future frames starting from the two input frames.

\end{enumerate}

\section{Algorithms}

\noindent\textbf{Single Sequence}
Algorithm~\ref{alg:single} provides the training steps of the \textit{single sequence} experiments.
\algrenewcommand\alglinenumber[1]{} 
\begin{algorithm}
\small
\caption{Single Sequence Training}
\begin{algorithmic}[1]
\Require Images $I_t$, $t \in T$, NeRF model $F_{\Theta}$, latents $z \in \{z_{t_0}, z_{t_1}\}$, ODE-RNN, nODE model $f_{\theta}$, decoder $\mathcal{D}$.
\State Initialize latent codes $z_{t_0}, z_{t_1}$ for the first two frames
\While{\textit{warm-up} iterations not completed}
    \State $\hat{I} = F_{\Theta}(z)$ 
    \State $\mathcal{L} \gets \mathcal{L}_{\text{NeRF}}(\hat{I}_k, I_k),\ k \in \{0,1\}$
    \State $\Theta, z \gets \text{Update}()$
\EndWhile

\While{\textit{joint training} iterations not completed}
    \State $(\mu, \sigma) \gets \text{ODE-RNN}(z_{t_0}, z_{t_1})$
    \State $z_{t_0}^{\text{dyn}} \sim \mathcal{N}(\mu, \sigma)$ 
    \State $z_{t_i}^{\text{dyn}} = \text{ODESolve}(f_{\theta},z_{t_0}^{\text{dyn}} ,t_i)$
    \State $z_{t_i} = \mathcal{D}(z_{t_i}^{\text{dyn}})$
    \State $\hat{I}_i = F_{\Theta}(z_{t_i})$ 
    \State $\mathcal{L} \gets \mathcal{L}_{\text{NeRF}}(\hat{I}_i, I_i)$
    \State $\Theta, \text{ODE-RNN}, f_{\theta}, \mathcal{D} \gets \text{Update}()$

\EndWhile

\end{algorithmic}
\label{alg:single}
\end{algorithm}

\noindent\textbf{Multi Sequence Generalization}
Algorithm~\ref{alg:multi} provides the training steps of the \textit{multi sequence generalization} experiments
\algrenewcommand\alglinenumber[1]{} 
\begin{algorithm}
\small
\caption{Generalized Multi-Sequence Training}
\begin{algorithmic}[1]
\Require Images $I_t^c$, $t \in T$ of sequence $c \in \{ 0, \ldots, N \}$, static background $I_{\text{static}}$, initial poses $p^c_0$, velocities $v^c_0$. Poses $p^c_{t_i}$, velocities $v^c_{t_i}$ to predict, NeRF model $F_{\Theta}$, latents $z_{\text{static}}, z_{\text{can}}$, encoder $\mathcal{E}$, nODE model $f_{\theta}$, decoders for velocity, pose, scene: $\mathcal{D}_v$, $\mathcal{D}_p$, $\mathcal{D}_n$.
\State Initialize $z_{static}$ for the static background
\While{\textit{warm-up} iterations not completed}
    \State $\hat{I} = F_{\Theta}(z_{static})$ 
    \State $\mathcal{L} \gets \mathcal{L}_{\text{NeRF}}(\hat{I}_{static}, I_{static})$
    \State $\Theta, z_{static} \gets \text{Update}()$
\EndWhile
\While{\textit{joint training} iterations not completed}
    \State $z_{t_0,c}^{\text{dyn}} = \text{concat}\left( z_{\text{can}}, \mathcal{E}\left(p^c_0\right), v^c_0 \right)$; 
    \State $z_{t_i,c}^{\text{dyn}} = \text{ODESolve}(f_{\theta}, z_{t_0,c}^{\text{dyn}}, t_i)$
    \State $z_{t_i,c} = \mathcal{D}_n(z_{t_i,c}^{\text{dyn}}) + z_{static}$
    \State $\hat{p}^c_{t_i}, \hat{v}^c_{t_i}, \hat{I}^c_{t_i} \gets \mathcal{D}_p(z_{t_i,c}^{\text{dyn}}), \mathcal{D}_v(z_{t_i,c}^{\text{dyn}}), F_{\Theta}(z_{t_i,c})$
    \State $\mathcal{L} \gets \mathcal{L}_{\text{NeRF}}(\hat{I}^c_{t_i}, I^c_{t_i})$\\
        $\qquad\qquad + \mathcal{L}_{\text{p}}(\hat{p}^c_{t_i}, p^c_{t_i}) + \mathcal{L}_{\text{v}}(\hat{v}^c_{t_i}, v^c_{t_i}) + \mathcal{L}_{\text{lipschitz}}$
    \State $\Theta, f_{\theta}, \mathcal{D}_n, \mathcal{D}_p, \mathcal{D}_v, z_{can} \gets \text{Update}()$
\EndWhile
\end{algorithmic}
\label{alg:multi}
\end{algorithm}

\section{Analysis}
\subsection{Effects of the number of nODE layers}
Table ~\ref{tab:ablation_node_layers} presents the ablation study on the \textit{Oscillating Ball} dataset, analyzing the effect of the number of layers in the nODE MLP $f_\theta$. Each experiment runs for 400k iterations. We observe a performance drop in the 5-layer $f_\theta$, where the model struggles to generate sequences with novel initial conditions. Although the 7-layer $f_\theta$ shows an improvement in performance compared to the 5-layer variant, it still underperforms relative to the 3-layer model.
We also observe from the visual results that, on the \textit{Bifurcating Hill} dataset, the 7-layered \( f_\theta \) completely underfits the scene and fails to learn the dynamics, whereas the 5-layered \( f_\theta \) captures the dynamics but underperforms compared to the 3-layered \( f_\theta \).

\begin{table}[ht]
\caption{Ablation study of nODE layers on different evaluation metrics in \textit{Oscillating Ball} dataset.}
    \vspace{-10pt}
    \small
    \centering
    \begin{tabular}{c|c c c c}
        \toprule
        {\# nODE Layers} & {SSIM} & {LPIPS} & {PSNR} & {IoU} \\
        \midrule
        3 & 0.662 & 0.4364 & 29.091 & 0.3327   \\
        5 & 0.605 & 0.5171 & 27.580 & 0.1864 \\
        7 & 0.645 & 0.4721 & 28.858 &  0.3060 \\
        \bottomrule
    \end{tabular}
    \label{tab:ablation_node_layers}
\end{table}

\subsection{Effects of the number of wamrup latents}
Table~\ref{tab:ablation_warmup_latent_count} presents an ablation study on the number of latents trained during the warmup stage in the single-sequence setup. The evaluation is conducted on the \textit{reconstruction} split of the \textit{Bouncing Balls} dataset. Quantitatively, the results remain largely comparable across different configurations. However, qualitative observations indicate that using a higher number of warmup latents leads to more stable long-term extrapolations. We select two warmup latents as it represents the minimum number required to capture the initial dynamics of the scene.

\begin{table}[ht]
\caption{Ablation studies of warmup latent count on \textit{Reconstruction} set of \textit{Bouncing Balls}}
    \vspace{-10pt}
    \small
    \centering
    \begin{tabular}{c|c c c c}
        \toprule
        {\# Count} & {SSIM} & {LPIPS} & {PSNR}  \\
        \midrule
        2 & 0.978 & 0.0310 & 33.70   \\
        5 & 0.978 & 0.0289 & 32.90  \\
        10 & 0.979 & 0.0307 & 33.54  \\
        \bottomrule
    \end{tabular}
    \label{tab:ablation_warmup_latent_count}
\end{table}

\subsection{Training Time}
Table~\ref{tab:training_time} reports the training time for each scene. The generalization experiments involving multiple sequences take significantly longer than the single-sequence experiments, requiring roughly three days to complete. Moreover, the training time grows proportionally with the length of the input sequences, highlighting a current limitation of our method.

\begin{table}[H]
\centering
\begin{tabular}{lcc}
\toprule
\textbf{Dataset} & \textbf{Solver} & \textbf{Training Time} \\
\midrule
Bouncing Balls & \texttt{dopri5} & 18 hours \\
Pendulum       & \texttt{euler}  & 24 hours \\
Sear Steak  & \texttt{euler} & 36 hours \\
Oscillating Ball  & \texttt{euler} & 72 hours \\
Bifurcating Hill  & \texttt{euler} & 72 hours \\
\bottomrule
\end{tabular}
\caption{Training time per dataset.}
\label{tab:training_time}
\end{table}


\end{document}